\documentclass[letterpaper]{article} 
\usepackage{aaai2026}  
\usepackage{times}  
\usepackage{helvet}  
\usepackage{courier}  
\usepackage[hyphens]{url}  
\usepackage{graphicx} 
\usepackage{natbib}  
\usepackage{caption} 
\usepackage{algorithm}
\usepackage{algorithmic}

\usepackage{amsmath}
\usepackage{amsfonts}
\usepackage{booktabs}
\usepackage{multirow}
\usepackage[table]{xcolor}
\usepackage{subcaption} %

\usepackage{newfloat}
\usepackage{listings}
\DeclareCaptionStyle{ruled}{labelfont=normalfont,labelsep=colon,strut=off} 
\floatstyle{ruled}
\newfloat{listing}{tb}{lst}{}
\floatname{listing}{Listing}
\title{Simba: Towards High-Fidelity and Geometrically-Consistent Point Cloud Completion via Transformation Diffusion}
\author{
    Lirui Zhang\textsuperscript{\rm 1}\equalcontrib,
    Zhengkai Zhao\textsuperscript{\rm 1}\equalcontrib,
    Zhi Zuo\textsuperscript{\rm 1},
    Pan Gao\textsuperscript{\rm 1}\thanks{Corresponding authors.},
    Jie Qin\textsuperscript{\rm 1}\footnotemark[2]
}
\affiliations{
    \textsuperscript{\rm 1}Nanjing University of Aeronautics and Astronautics\\
    Nanjing, Jiangsu, China\\
    \{lirui.zhang, zhengkai.zhao, zuozhi, pan.gao, jie.qin\}@nuaa.edu.cn
}

\usepackage{bibentry}
\begin{document}
\maketitle

\begin{abstract}
Point cloud completion is a fundamental task in 3D vision. A persistent challenge in this field is simultaneously preserving fine-grained details present in the input while ensuring the global structural integrity of the completed shape. While recent works leveraging local symmetry transformations via direct regression have significantly improved the preservation of geometric structure details, these methods suffer from two major limitations: 
(1) These regression-based methods are prone to overfitting which tend to memorize instant-specific transformations instead of learning a generalizable geometric prior. (2) Their reliance on point-wise transformation  regression lead to high sensitivity to input noise, severely degrading their robustness and generalization.
To address these challenges, we introduce \textbf{Simba}, a novel framework that reformulates point-wise transformation regression as a distribution learning problem. Our approach integrates symmetry priors with the powerful generative capabilities of diffusion models, avoiding instance-specific memorization while capturing robust geometric structures. 
Additionally, we introduce a hierarchical Mamba-based architecture to achieve high-fidelity upsampling. Extensive experiments across the PCN, ShapeNet, and KITTI benchmarks validate our method's state-of-the-art (SOTA) performance. Codes are available at \url{https://github.com/I2-Multimedia-Lab/Simba}.

\end{abstract}

\section{Introduction}
Point clouds, as a fundamental 3D representation, are integral to numerous applications, from autonomous driving to robotics and augmented reality~\cite{geiger2012we,cadena2017past, proxyformer,zuo2025uni4dunifiedselfsupervisedlearning}. Point clouds captured in real-world environments are often incomplete due to factors such as occlusion and limited sensor range, making point cloud completion a critical area of research~\cite{pcn,10261217,li2024global}.

\begin{figure}[t]
    \centering

    \begin{subfigure}[b]{\linewidth}
        \centering
        \includegraphics[width=1\linewidth]{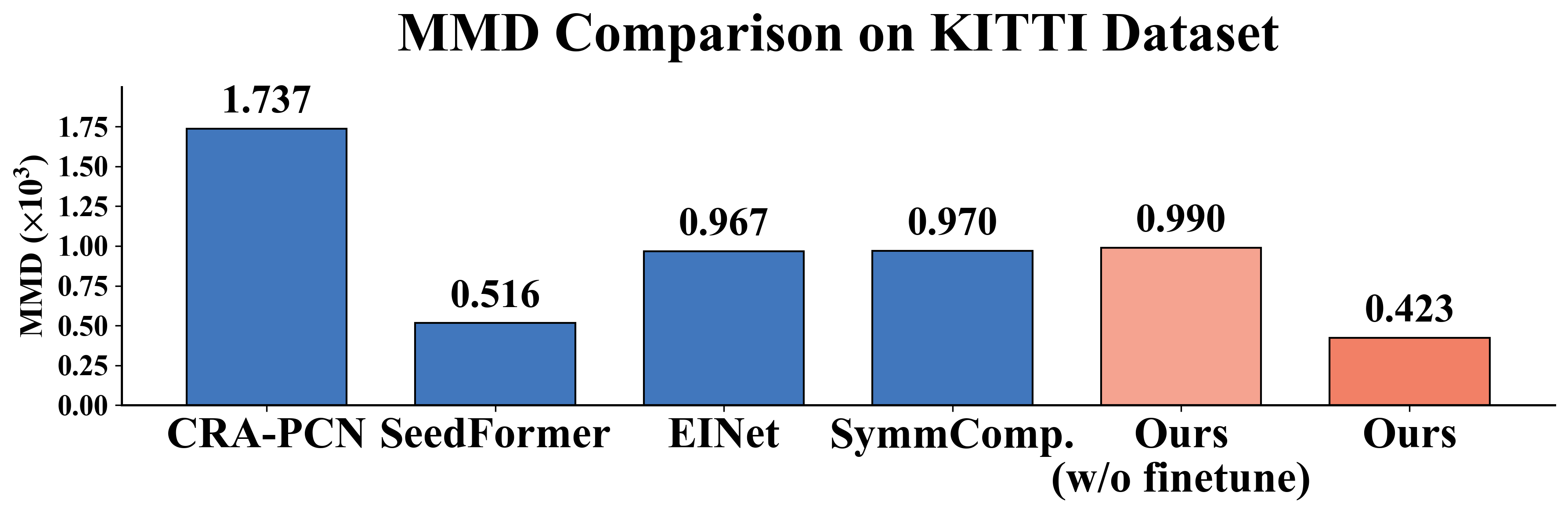}
    \end{subfigure}

    \begin{subfigure}[b]{\linewidth}
        \centering
        \includegraphics[width=1\linewidth]{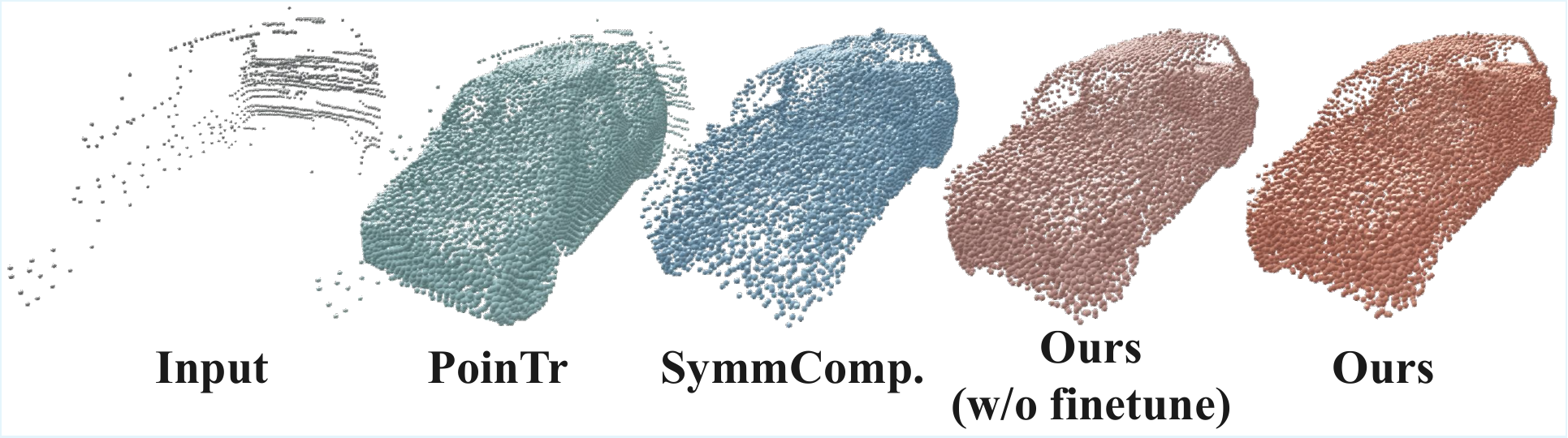}
    \end{subfigure}

    \caption{
    Strong cross-domain generalizability on KITTI. Our model, trained on synthetic data, is competitive without finetuning and achieves superior performance with it.
    }
    \label{fig:kitti_visuals}
\end{figure}
Existing completion methods often struggle to preserve fine-grained details from the input while maintaining the global structural integrity of the completed shape~\cite{pcn,wen2021pmp,grnet}. Leveraging symmetry to enforce structural consistency has emerged as a powerful approach, and the explicit modeling of geometric priors has shown great potential~\cite{schiebener2016opening,cui2023geometric}. However, existing approaches to modeling symmetry have notable limitations: On one hand, many methods rely on the assumption that objects are globally axisymmetric, which limits their ability to capture local or partial symmetries~\cite{zhang2023GTNet,ma2023symmetry}. Recent work like SymmCompletion~\cite{symmcompletion} has shown the promise of learning point-wise local symmetry transformations, but its reliance on \textbf{direct regression} is a critical limitation: (1) they tend to overfit the training distribution, memorizing specific transformation patterns instead of learning generalized geometric alignment rules; (2) they are highly sensitive to occlusions and noise, as each point is treated independently, resulting in fragmented or distorted global structures. These limitations severely hinder generalization to real-world or unseen data distributions~\cite{geiger2012we}.

How to leverage the powerful symmetry priors encoded in affine and translation matrices while avoiding the risk of the network merely memorizing specific transformation patterns remains an open problem. We observe that diffusion models possess strong generative capabilities, enabling diverse sampling. By integrating diffusion with transformation matrices, we can effectively utilize geometric priors without overfitting to fixed solutions.

To this end, we propose a novel direction that leverages the generative power of diffusion models. We argue that prior applications of diffusion, which directly operate on point coordinates \cite{pdr}, are suboptimal as they can wash out fine details present in the partial input. To explore symmetric geometric priors completely, we reframe the completion task and propose a novel generative paradigm: \textbf{instead of diffusing points, we diffuse a field of geometric transformations}. Our framework, Simba, is the first to learn the conditional distribution of point-wise affine transformations. By iteratively denoising a low-dimensional transformation vector, our model generates a robust and geometrically plausible structural prior, which, when applied to the original keypoints, constructs a complete shape while inherently preserving its details. As shown in Figure~\ref{fig:kitti_visuals}, our model, both with and without fine-tuning, outperforms competing baselines on the real-world KITTI vehicle completion task. This achievement underscores the strong cross-domain generalizability of our approach and its powerful synthetic-to-real transfer capabilities.

Specifically, we propose a two-stage learning framework. In Stage 1, we pre-train SymmGT, a supervision network that generates target transformation matrices for the subsequent stage. In Stage 2, we employ a diffusion model, termed Symmetry-Diffusion (Sym-Diffuser), which conditions on partial input features to generate transformation fields. This design fully exploits the generative capacity of diffusion models to capture symmetric geometric priors. Subsequently, a Mamba-based refinement (MBA-Refiner) network is introduced to progressively enhance and upsample the coarse completions into high-fidelity outputs.

Extensive experiments on multiple benchmarks demonstrate that our method achieves state-of-the-art (SOTA) performance. In addition, we conduct evaluations on the real-world KITTI dataset, further validating the generalizability and effectiveness of our approach.
Our main contributions are summarized as follows:
\begin{itemize}
    \item We propose \textbf{\textit{Simba}}, a novel framework that formulates point cloud completion as a conditional generative task over a field of geometric transformations.
    \item We are the \textbf{\textit{first}} to employ a diffusion model, termed \textbf{\textit{Sym-Diffuser}}, to learn the distribution of affine transformations, providing a robust representation to ensure the geometric consistency of the completed shape.
    \item We design the \textbf{\textit{MBA-Refiner}}, a cascaded Mamba-based architecture, to progressively refine the coarse output, enabling high-fidelity progressive upsampling.
\end{itemize}
\section{Related Work}
\subsection{Point Cloud Completion}
Foundational works such as PointNet~\cite{pointnet} and PointNet++~\cite{pointnet++} enabled end-to-end learning on point clouds, laying the groundwork for point cloud completion networks. Early methods like PCN~\cite{pcn} and FoldingNet~\cite{foldingnet} adopted a canonical coarse-to-fine paradigm, learning global shape priors to generate complete surfaces. This approach was further refined by architectures such as SnowflakeNet~\cite{snowflakenet}, which incorporated sophisticated decoders to produce more uniform and detailed point distributions. With the advent of transformers~\cite{transformer}, recent methods~\cite{zhao2021point,proxyformer,rong2024cra,wang2024pointattn,nunes2024scaling,yu2024context} have focused on capturing long-range dependencies within point sets, achieving state-of-the-art performance in completion tasks. However, the quadratic complexity of transformers presents a major bottleneck in computational efficiency. To alleviate this issue, 3DMambaComplete~\cite{3dmambacomplete} introduced the Mamba~\cite{gu2023mamba,liang2024pointmamba} architecture into point cloud completion, significantly reducing computational overhead.

Diffusion models~\cite{ddpm,ddim} have inspired several novel paradigms for 3D synthesis. These include directly denoising 3D coordinates, as in PDR~\cite{pdr}, leveraging 2D priors through optimization (SDS)~\cite{sds}, or fusing outputs from feed-forward image-to-3D models like PCDreamer~\cite{pcdreamer,Li_2025_CVPR}. Despite their impressive generative capabilities, such approaches often suffer from high computational costs, slow inference, and difficulty in faithfully preserving input details during fusion.
\begin{figure}[t]
    \centering
    \includegraphics[width=0.48\textwidth]{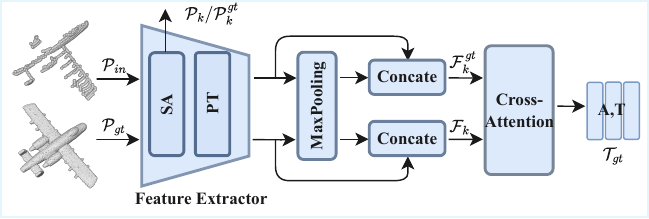}
    \caption{
        \textbf{Stage 1: SymmGT pre-training architecture.} 
        The network regresses transformation field $\boldsymbol{\mathcal{T}}_{gt}$ from partial input and complete GT point clouds.
    }
    \label{fig:stage1}
\end{figure}

\subsection{Symmetry Priors in Point Cloud Completion}
Symmetry serves as a fundamental geometric prior for completing structured objects, particularly man-made shapes. Early approaches like GTNet~\cite{gtnet} relied on global symmetry assumptions, limiting their effectiveness on partial or asymmetrical inputs. SymmCompletion~\cite{symmcompletion} advanced this by learning point-wise local affine transformations between observed and missing regions. However, deterministic transformation regression from partial observations remains ill-posed and prone to overfitting on training-specific patterns.

In contrast, our method formulates point-wise transformation prediction as a distribution learning problem, leveraging the intrinsic diversity of diffusion models to mitigate overfitting. Furthermore, we introduce a cascaded Mamba-based refinement network block combined with a hierarchical upsampling strategy to enhance computational efficiency without sacrificing performance.

\begin{figure*}[t]
    \centering
    \includegraphics[width=0.95\linewidth]{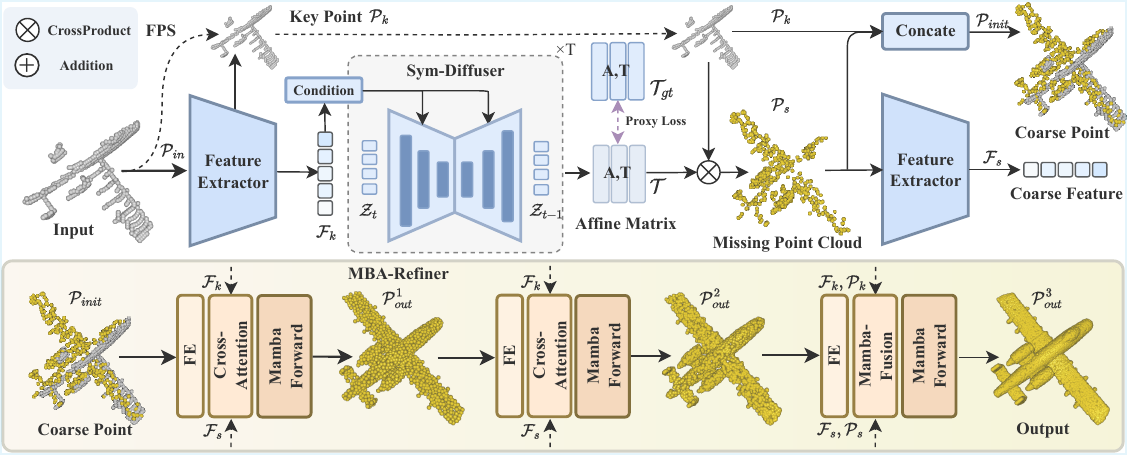}
    \caption{
            \textbf{Stage 2: The Simba coarse-to-fine architecture.} 
            Our framework comprises two core components: a Symmetry-Diffusion Module \textbf{(Sym-Diffuser)} that generates a field of geometric transformations to produce a structurally-complete coarse shape from partial input, and a novel \textbf{MBA-Refiner} decoder that progressively refines this coarse representation to yield the final high-fidelity output point cloud. \textbf{FE} denotes Feature Extractor.
        }
    \label{fig:pointcloud_architecture}
\end{figure*}
\section{Methodology}

The two-stage architecture of our Simba framework is illustrated in Figures~\ref{fig:stage1} and~\ref{fig:pointcloud_architecture}. Stage 1 (Fig~\ref{fig:stage1}) pre-trains a \textbf{SymmGT} network to generate target transformation matrices for diffusion supervision. Stage 2 (Fig~\ref{fig:pointcloud_architecture}) comprises two core components: a Symmetry-Diffusion Module \textbf{(Sym-Diffuser)} and a cascaded Mamba-Based Refinement network \textbf{(MBA-Refiner)}. The Sym-Diffuser produces a point-wise affine transformation field to generate a coarse yet structurally complete point cloud from partial input. The MBA-Refiner progressively refines and upsamples this coarse prediction through cascaded processing to synthesize the final high-fidelity output.

\subsection{Pre-training SymmGT (Stage 1)}
We observe that symmetric geometric priors inherently capture strong geometric awareness, which can serve as an effective inductive bias for shape completion. Inspired by~\cite{symmcompletion}, we incorporate such priors into our generative framework. In this section, our target is to generate the transformation matrix that guides the diffusion process. 
Specifically, we pre-train a base network, SymmGT. As shown in Figure~\ref{fig:stage1}, given a partial input point cloud $\boldsymbol{\mathcal{P}}_{in}$ and a complete ground truth (GT) point cloud $\boldsymbol{\mathcal{P}}_{gt}$, it first samples a set of keypoints $\boldsymbol{\mathcal{P}}_k$ from $\boldsymbol{\mathcal{P}}_{in}$ and extracts the keypoint features $\boldsymbol{\mathcal{F}}_k$ from the partial input and a global feature $\boldsymbol{\mathcal{F}}_{gt}$ from the GT point cloud by a \textbf{shared-weight Feature Extractor}, which consists of a Set Abstraction (SA) layer~\cite{pointnet++} and a Point Transformer block~\cite{zhao2021point}. These features are fused via cross-attention to regress a target transformation field, denoted as $\boldsymbol{\mathcal{T}}_{gt} \in \mathbb{R}^{K \times 12}$. This field is composed of a point-wise affine matrix $\boldsymbol{A}_i \in \mathbb{R}^{3 \times 3}$ and a translation vector $\boldsymbol{T}_i \in \mathbb{R}^{3}$. The transformation field $\boldsymbol{\mathcal{T}}$ is then applied to the input keypoints $\boldsymbol{\mathcal{P}}_k$ to construct a coarse but complete point cloud $\boldsymbol{\mathcal{P}}_{init}$. The network is trained by minimizing the Chamfer Distance ($L_{CD}$) between this reconstructed coarse shape and the ground truth:
\begin{equation}
    \mathcal{L}_{\text{stage1}} = L_{CD}(\boldsymbol{\mathcal{P}}_{k} \cup \{ \boldsymbol{A}_i \mathbf{p}_i + \boldsymbol{T}_i \}, \boldsymbol{\mathcal{P}}_{gt}),
    \label{eq:stage1_loss}
\end{equation}
where $(\boldsymbol{A}_i, \boldsymbol{T}_i) \in \boldsymbol{\mathcal{T}}$, $\mathbf{p}_i \in \boldsymbol{\mathcal{P}}_k$, and $k$ denotes the keypoint set. In Stage 2, the frozen SymmGT is used solely to generate this target field $\boldsymbol{\mathcal{T}}{gt}$, which serves as the clean data target $\mathcal{Z}_0$ for training our Sym-Diffuser.
\subsection{Completion with Simba (Stage 2)}

While SymmCompletion ~\cite{symmcompletion} shows notable improvements on datasets like PCN and ShapeNet, its performance on real-world data remains limited. We attribute this to the inherent limitations of its regression-based approach, namely overfitting and high sensitivity to occlusions and noise. To address these fundamental issues, we propose a novel generative paradigm: instead of diffusing points, we diffuse a field of geometric transformations by leveraging the generative power of Diffusion Models.

\begin{figure*}[t!]
    \centering
    \includegraphics[width=0.93\textwidth]{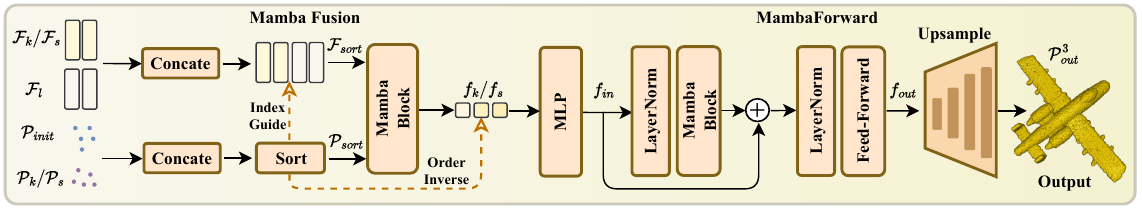}
    \caption{
        \textbf{Detailed architecture of the Mamba Fusion and MambaForward module.} 
        It employs a \textbf{Mamba Fusion} block to coherently merge multi-source features and point coordinates. 
        A core \textbf{MambaForward}, integrated within a feed-forward network, then progressively refines and upsamples the geometry to produce a high-fidelity output.
    }
    \label{fig:Mamba_Fusion}
\end{figure*}
\subsubsection{Symmetry-Diffusion Module (Sym-Diffuser)}
\label{sec:Sym-Diffuser}
The Sym-Diffuser is designed to generate a high-quality, structurally sound coarse completion by producing a field of symmetric transformations. Given an input point cloud $\boldsymbol{\mathcal{P}}_{in}$, we employ a conditional diffusion model, Sym-Diffuser, to generate its corresponding transformation field. The model learns the conditional distribution $p(\boldsymbol{\mathcal{T}} | \boldsymbol{\mathcal{F}}_k)$, where $\boldsymbol{\mathcal{F}}_k$ are features extracted from the input keypoints.

During training, we leverage the ground truth transformation field $\boldsymbol{\mathcal{T}}_{gt}$ from Stage~1 as clean data, denoted $\boldsymbol{\mathcal{Z}}_0$. We simulate the standard forward diffusion process, a sequential Markov chain that progressively corrupts $\boldsymbol{\mathcal{Z}}_0$ with Gaussian noise over $T=100$ timesteps. This process creates a series of noisy versions $\boldsymbol{\mathcal{Z}}_t$, where the noise level at any timestep $t$ is governed by a predefined variance schedule. This allows for efficient, direct sampling of any noisy version $\boldsymbol{\mathcal{Z}}_t$ from $\boldsymbol{\mathcal{Z}}_0$ via a closed-form expression.

Our diffusion model is trained to reverse this process. Internally, it functions as a noise predictor $\boldsymbol{\epsilon}_{\theta}$ that estimates the noise in a given input $\boldsymbol{\mathcal{Z}}_t$. This allows us to recover the predicted clean field $\hat{\boldsymbol{\mathcal{T}}}_{\theta}$ by algebraically inverting the forward process with the predicted noise. Inspired by Consistency Models~\cite{proxyloss}, our training objective is formulated as a weighted Mean Squared Error (MSE) between the predicted clean field $\hat{\boldsymbol{\mathcal{T}}}_{\theta}$ and the ground truth $\boldsymbol{\mathcal{T}}_{gt}$. A weighting function $\lambda(t)$ modulates the contribution of each timestep, prioritizing learning at different signal-to-noise ratios.
\begin{equation}
    \mathcal{L}_{\text{proxy}} = \mathbb{E}_{t, \boldsymbol{\mathcal{Z}}_0, \boldsymbol{\epsilon}} \left[ \lambda(t) \left\| \boldsymbol{\mathcal{T}}_{gt} - \hat{\boldsymbol{\mathcal{T}}}_{\theta}(\boldsymbol{\mathcal{Z}}_t, t, \boldsymbol{\mathcal{F}}_k) \right\|^2 \right]
    \label{eq:proxy_loss}
\end{equation}

At inference, the Sym-Diffuser takes a random Gaussian vector $\boldsymbol{Z} \in \mathbb{R}^{N_k \times 12}$ and, conditioned on the features $\boldsymbol{\mathcal{F}}_k$, iteratively denoises it to produce a clean transformation field $\hat{\boldsymbol{\mathcal{T}}} \in \mathbb{R}^{N_k \times 12}$. Following~\cite{symmcompletion}, this field is composed of a point-wise affine matrix $\boldsymbol{A}_i \in \mathbb{R}^{3 \times 3}$ and a translation vector $\boldsymbol{T}_i \in \mathbb{R}^{3}$ for each keypoint $\mathbf{p}_i \in \boldsymbol{\mathcal{P}}_k$. The symmetric keypoints $\boldsymbol{\mathcal{P}}_s$ are then constructed by applying these transformations as follows:
\begin{equation}
    \boldsymbol{\mathcal{P}}_s = \{ \boldsymbol{A}_i \mathbf{p}_i + \boldsymbol{T}_i \mid \mathbf{p}_i \in \boldsymbol{\mathcal{P}}_{k}, (\boldsymbol{A}_i, \boldsymbol{T}_i) \in \hat{\boldsymbol{\mathcal{T}}} \}
    \label{eq:transformation_apply}
\end{equation}

The initial coarse completion is formed by the union of the partial keypoints and the generated missing part, $\boldsymbol{\mathcal{P}}_{init} = \boldsymbol{\mathcal{P}}_{k} \cup \boldsymbol{\mathcal{P}}_{s} \in \mathbb{R}^{2N_k \times 3}$.

\subsubsection{Cascaded Refinement with MBA-Refiner}
\label{sec:MBA-Refiner}
The coarse completion $\boldsymbol{\mathcal{P}}_{init}$, encoded by FE to $\boldsymbol{\mathcal{F}}_{l}$, is refined and upsampled by the MBA-Refiner, a three-block cascade balancing performance and efficiency. Each block follows a consistent design: Feature Fusion to integrate guidance, followed by MambaForward for refinement and upsampling. The blocks differ in their fusion strategy, tailored to point density. The refinement is guided by partial keypoint features $\boldsymbol{\mathcal{F}}_k$ and symmetric point features $\boldsymbol{\mathcal{F}}_s$.

The MBA-Refiner employs different feature fusion strategies across its three blocks to adapt to varying computational constraints at different point densities.

\textbf{Blocks 1-2: Cross-Attention Fusion.} At lower point densities ($l = 0, 1$), we employ cross-attention fusion for performance. The base features $\boldsymbol{\mathcal{F}}_{l}$ from the previous layer are refined by separately attending to each guidance source via Multi-head Cross-Attention (MCA). Specifically, $\boldsymbol{\mathcal{F}}_{l}$ attends to $\boldsymbol{\mathcal{F}}_k$ and $\boldsymbol{\mathcal{F}}_s$ independently, and the resulting context-aware features are concatenated and processed through a Multi-Layer Perceptron (MLP), denoted as a fusion function $\boldsymbol{\psi}$, to produce the unified feature set $\boldsymbol{f}_{in}^{l}$ for subsequent refinement, where $[\cdot]$ denotes the concatenation operation:
\begin{equation}
    \boldsymbol{f}_{in}^{l} = \boldsymbol{\psi} \left( \big[ \text{MCA}(\boldsymbol{\mathcal{F}}_{l}, \boldsymbol{\mathcal{F}}_g) \big]_{g \in \{k, s\}} \right)
    \label{eq:mca_fusion}
\end{equation}

\textbf{Block 3: Mamba Fusion.} At the highest point density ($l = 2$), where the $\mathcal{O}(N^2)$ complexity of attention is prohibitive, we employ an efficient Mamba-based fusion (MFusion) strategy as illustrated in Figure~\ref{fig:Mamba_Fusion}. The base features $\boldsymbol{\mathcal{F}}_{l}$ from the previous layer are fused with guidance features ($\boldsymbol{\mathcal{F}}_k$ and $\boldsymbol{\mathcal{F}}_s$) through spatial ordering and Mamba processing, then processed to produce the unified feature set $\boldsymbol{f}_{in}^{3}$ for subsequent refinement:
\begin{equation}
    \boldsymbol{f}_{in}^{3} = \boldsymbol{\psi} \left( \big[ \text{MFusion}(\boldsymbol{\mathcal{F}}_{l}, \boldsymbol{\mathcal{F}}_g) \big]_{g \in \{k, s\}} \right)
\end{equation}

\textbf{Shared Refinement with MambaForward.}
After the fusion stage, all blocks utilize a shared-architecture \textbf{MambaForward} module for final feature refinement and upsampling. As shown in Figure~\ref{fig:Mamba_Fusion}, this module takes the fused features from the corresponding layer, $\boldsymbol{f}_{in}^{l}$, and processes them through a sequence of operations including an MLP, a Mamba block with residual connections, and an upsampling layer to directly produce the refined point cloud:
\begin{equation}
    \boldsymbol{\mathcal{P}}_{out}^{l} = \text{MambaForward}(\boldsymbol{f}_{in}^{l})
\end{equation}
This shared design ensures consistent and powerful refinement across the cascade, achieving $2\times$ upsampling in Blocks 1-2 and $4\times$ in Block 3.
\begin{table*}[t]
    \centering

    \scalebox{0.85}{
    \begin{tabular}{l|c|cccccccc|c}
        \toprule
        \textbf{Methods} & \textbf{\cellcolor{white}Avg CD-$\ell_1$} & \textbf{Plane} & \textbf{Cabinet} & \textbf{Car} & \textbf{Chair} & \textbf{Lamp} & \textbf{Sofa} & \textbf{Table} & \textbf{Watercraft} & \textbf{F-Score@1\%} \\
        \midrule
        PCN(3DV 2018) & 9.64 & 5.50 & 22.70 & 10.63 & 8.70 & 11.00 & 11.34 & 11.68 & 8.59 & 0.695 \\
        PoinTr(ICCV2021) & 8.38 & 4.75 & 10.47 & 8.68 & 9.39 & 7.75 & 10.93 & 7.78 & 7.29 & - \\
        SnowflakeNet(ICCV2021) & 7.21 & 4.29 & 9.16 & 8.08 & 7.89 & 6.07 & 9.23 & 6.55 & 6.40 & 0.801 \\
        SeedFormer(ECCV2022) & 6.74 & 3.85 & 9.05 & 8.06 & 7.06 & 5.21 & 8.85 & 6.05 & 5.85 & 0.818 \\
        ProxyFormer(CVPR2023) & 6.77 & 4.01 & 9.01 & 7.88 & 7.11 & 5.35 & 8.77 & 6.03 & 5.98 & - \\
        AdaPoinTr(TPAMI2023) & 6.53 & 3.68 & 8.82 & 7.47 & 6.85 & 5.47 & 8.35 & 5.80 & 5.76 & 0.845 \\
        SVDFormer(ICCV2023) & 6.54 & 3.62 & 8.79 & 7.46 & 6.91 & 5.33 & 8.49 & 5.90 & 5.83 & 0.841 \\
        CRA-PCN(AAAI2024) & 6.39 &  3.59 & 8.70 & 7.50 & 6.70 & \textbf{5.06} & 8.24 & 5.72 & 5.64 & - \\
        SymmCompletion(AAAI2025)  & 6.47 & 3.67 & 8.74 & 7.47 & 6.86 & 5.11 & 8.41 & 5.88 & 5.66 & 0.840 \\
        PointCFormer(AAAI2025) & 6.41 & 3.53 & 8.73 & 7.32 & \textbf{6.68} & 5.12 & 8.34 & 5.86 & 5.74 & 0.855 \\
        DC-PCN(AAAI2025) & 6.46 & 3.65 & 8.75 & 7.48 & 6.71 & 5.35 & 8.28 &  5.76 & 5.71 & 0.850 \\
        PCDreamer(CVPR2025) & 6.52 & \textbf{3.51} & \textbf{8.62} & \textbf{6.92} & 6.91 & 5.66 & 8.31 & 6.27 & 5.90 & \textbf{0.856} \\
        \midrule
        \textbf{Ours (Simba)} & \textbf{6.34} & 3.63 & 8.67 & 7.34 & 6.74 & 5.09 & \textbf{8.17} & \textbf{5.62} & \textbf{5.51} & 0.853 \\
        \bottomrule
    \end{tabular}}
    \caption{
        Quantitative comparison on the PCN dataset. We report per-category Chamfer Distance ($\ell_1$ CD $\times 10^3 \downarrow$) and average F-Score@1\% $\uparrow$. 
        Specifically, \textbf{Avg CD-$\ell_1$} (average $\ell_1$ Chamfer Distance) reflects reconstruction accuracy; 
        \textbf{F-Score@1\%} measures the geometric similarity. Results are from original papers or carefully reproduced under identical experimental settings.
    }
    \label{tab:pcn_results}
\end{table*}

\begin{figure*}[ht!]
    \centering
    \includegraphics[width=0.95\linewidth]{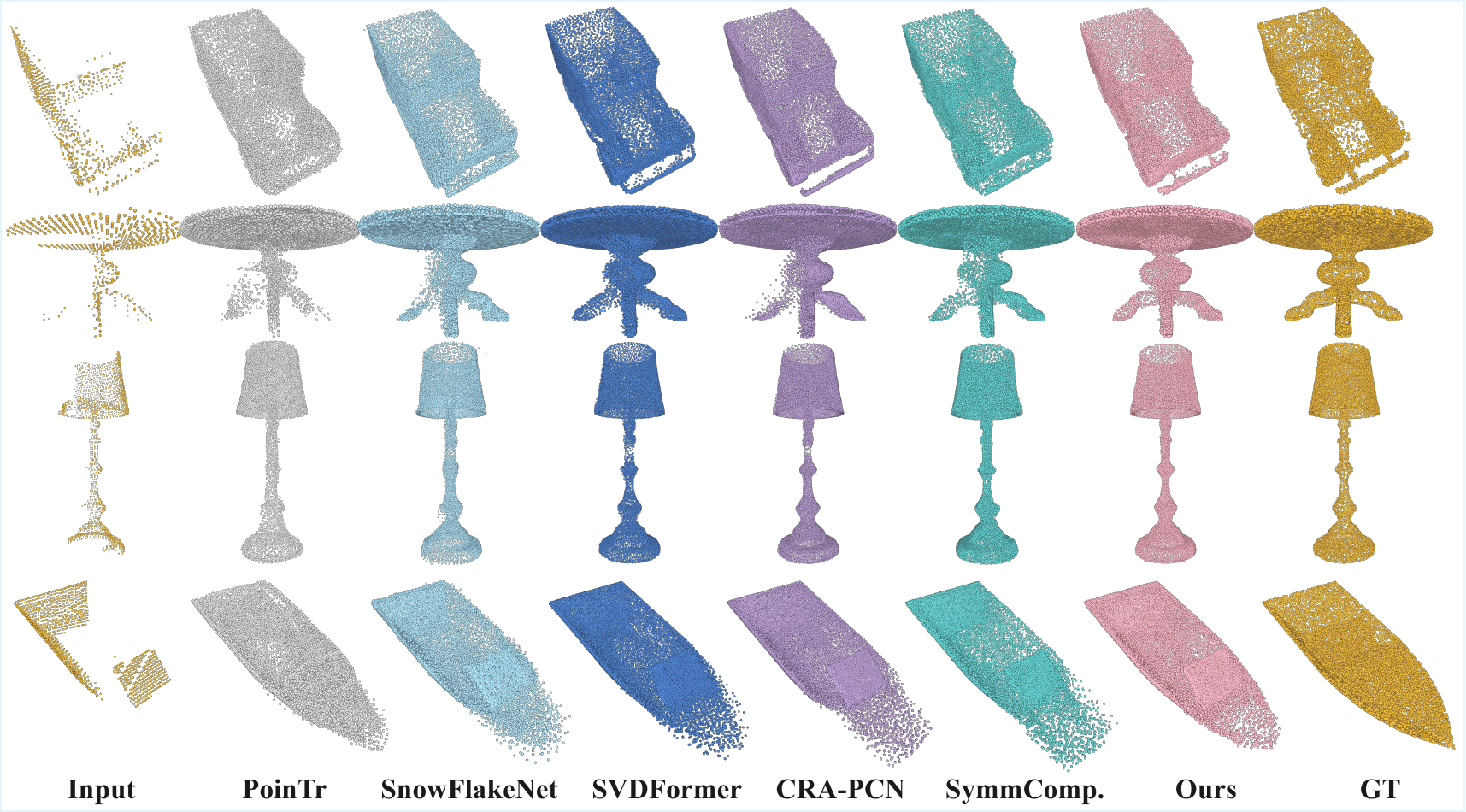}
    \caption{Qualitative comparison on the PCN dataset. Our \textbf{Simba} achieves superior geometric consistency and preserves fine details where other methods struggle.}
    \label{fig:pcn_visual_comparison}
\end{figure*}

\subsection{Overall Training Objective}
\label{sec:loss}
Our Simba framework employs a two-stage training strategy. In Stage 1, SymmGT is pre-trained using Chamfer Distance loss to generate target transformation fields. In Stage 2, the complete framework is trained end-to-end with a composite objective that supervises both Sym-Diffuser and MBA-Refiner simultaneously:
\begin{equation}
    \mathcal{L}_{\text{stage2}} = \mathcal{L}_{\text{proxy}} + \sum_{l=1}^{3} L_{CD}(\boldsymbol{\mathcal{P}}^{l}_{out}, \boldsymbol{\mathcal{P}}_{gt})
    \label{eq:total_loss}
\end{equation}
where $\mathcal{L}_{\text{proxy}}$ supervises the Sym-Diffuser module as defined in Eq.~\ref{eq:proxy_loss}, and the Chamfer Distance terms supervise the MBA-Refiner cascade at each refinement stage $\boldsymbol{\mathcal{P}}_{out}^{l}$. This multi-stage supervision ensures effective joint optimization of both generative and refinement components.

\section{Experiments}
\begin{table*}[t]
\centering

\small
\setlength{\tabcolsep}{5pt}
\renewcommand{\arraystretch}{1.1}
\begin{tabular}{l|cccc|cccc|cccc}
\toprule
\multirow{2}{*}{\textbf{Method}} 
& \multicolumn{4}{c|}{\textbf{ShapeNet-55}} 
& \multicolumn{4}{c|}{\textbf{ShapeNet-34 (Seen)}} 
& \multicolumn{4}{c}{\textbf{ShapeNet-21 (Unseen)}} \\
\cmidrule(lr){2-5} \cmidrule(lr){6-9} \cmidrule(lr){10-13}
& CD-S & CD-M & CD-H & CD-Avg 
& CD-S & CD-M & CD-H & CD-Avg 
& CD-S & CD-M & CD-H & CD-Avg \\
\midrule
FoldingNet  & 2.67 & 2.66 & 4.05 & 3.12 & 1.86 & 1.81 & 3.38 & 2.35 & 2.76 & 2.74 & 5.36 & 3.62 \\
PCN         & 1.94 & 1.96 & 4.08 & 2.66 & 1.87 & 1.81 & 2.97 & 2.22 & 3.17 & 3.08 & 5.29 & 3.85 \\
PoinTr      & 0.67 & 1.05 & 2.02 & 1.25 & 0.76 & 1.05 & 1.88 & 1.23 & 1.04 & 1.67 & 3.44 & 2.05 \\
SeedFormer  & 0.50 & 0.77 & 1.49 & 0.92 & 0.48 & 0.70 & 1.30 & 0.83 & 0.61 & 1.07 & 2.35 & 1.34 \\
AdaPoinTr   & 0.49 & 0.69 & \textbf{1.24} & 0.81 & 0.48 & 0.63 & \textbf{1.07} & 0.73 & 0.61 & \textbf{0.96} & \textbf{2.11} & 1.23 \\
SVDFormer   & 0.48 & 0.70 & 1.30 & 0.83 & 0.46 & 0.65 & 1.13 & 0.75 & 0.61 & 1.05 & 2.19 & 1.28 \\
CRA-PCN     & 0.48 & 0.71 & 1.37 & 0.85 & 0.45 & 0.65 & 1.18 & 0.76 & 0.55 & 0.97 & 2.19 & 1.24 \\
\midrule
\textbf{Ours (Simba)} & 
\textbf{0.45} & \textbf{0.66} & 1.25 & \textbf{0.79} &
\textbf{0.43} & \textbf{0.59} & {1.08} & \textbf{0.70} &
\textbf{0.54} & 0.97 & 2.18 & \textbf{1.23} \\
\bottomrule
\end{tabular}
\caption{
Comparison of point cloud completion performance on ShapeNet-55, ShapeNet-34 (seen), and ShapeNet-21 (unseen) datasets.
All metrics report the L2 Chamfer Distance ($\times 10^3$); lower is better. 
Best performance in each block is marked in \textbf{bold}.
}

\label{tab:unified_shapenet_completion_results}
\end{table*}
\subsection*{Implementation Details}

Our framework is implemented in PyTorch, and all experiments were conducted on four NVIDIA RTX 4090 GPUs. 

\begin{table}[h]
  \centering
  \small 
  \setlength{\tabcolsep}{4pt} 

  \begin{tabular}{l|ccccc}
    \toprule
    \textbf{Metric} & crapcn & SeedFormer  & EINet & SymmComp. & \textbf{Ours} \\
    \midrule
    MMD  & 1.737 & 0.516 & 0.967  & 0.970 & \textbf{0.423} \\
    \bottomrule
  \end{tabular}
  \caption{Quantitative comparison on the KITTI dataset using MMD ($\times 10^3$). Lower is better.}
\label{tab:kitti_results}
\end{table}


\subsection{Datasets and Evaluation Metrics}

\noindent\textbf{Datasets.}
Our experiments are conducted on three widely-used public datasets, covering both synthetic and real-world scenarios.
\textbf{PCN}~\cite{pcn} is a classic benchmark for point cloud completion, consisting of 8 categories from ShapeNet~\cite{shapenet}. We follow its official train/validation/test split.
\textbf{ShapeNet-55/34}~\cite{pointr} is a large-scale, more diverse dataset derived from ShapeNet. We use it to evaluate the generalization capability of our model, reporting results on the full 55 categories, as well as on the challenging 34 seen and 21 unseen category split.
\textbf{KITTI}~\cite{kitti} provides real-world LiDAR scans of vehicles. We use this dataset to assess the robustness and generalization performance of our model on sparse, noisy, and partial data from a different domain.

\noindent\textbf{Evaluation Metrics.}
We employ standard metrics to quantitatively evaluate the completion quality.
For synthetic datasets (PCN and ShapeNet), we use \textbf{Chamfer Distance (CD)} and \textbf{F-Score}. We report L1-CD for PCN and L2-CD for ShapeNet, following common practice. For a more fine-grained analysis on ShapeNet, we also report the L2-CD on three difficulty levels: Simple (S), Moderate (M), and Hard (H). The F-Score (1\% threshold) measures surface reconstruction accuracy and is less sensitive to outliers.

\subsection{Results on PCN Dataset}
\noindent\textbf{Quantitative Analysis.}
We first evaluate our \textbf{Simba} on the widely used PCN benchmark~\cite{pcn}. Table~\ref{tab:pcn_results} presents the quantitative comparison against state-of-the-art methods in terms of L1 Chamfer Distance (CD) and F1-Score. 
Our method achieves state-of-the-art performance on several categories, such as ``Sofa'', ``Table'', and ``Watercraft''. In other categories, our accuracy is also competitive with existing leading approaches. Most importantly, we obtain the best overall performance in terms of average Chamfer Distance. Furthermore, our method demonstrates satisfactory precision on the F-Score@1\% metric.

Notably, our model demonstrates superior performance over SymmCompletion. We attribute this to our novel transformation diffusion paradigm, where the generative Sym-Diffuser produces more robust and plausible symmetric geometric priors than deterministic regression approaches.

\noindent\textbf{Qualitative Analysis.} Figure~\ref{fig:pcn_visual_comparison} provides a visual testament to \textbf{Simba's} performance, showcasing its capacity to generate completions with exceptional \textbf{geometric consistency} and \textbf{high fidelity}. For complex shapes like 'Table' and 'Car', our model faithfully reconstructs intricate features, such as distinct table legs and the car's smooth, complete body. The results are free from the distorted geometry and fragmented artifacts that are prevalent in outputs from CAR-PCN, PoinTr, and SVDFormer. Simba's proficiency is not limited to large-scale coherence; it also excels at preserving fine-grained details. It restores the 'Watercraft' with a pristine, continuous hull, unlike the noisy results from SnowFlakeNet, and accurately reproduces the 'Lamp's' ornate stem, a challenging feature where competing methods often resort to coarse approximations. This balanced performance validates our method's ability to generate completions that are both globally plausible and locally precise.

\subsection{Results on ShapeNet Datasets}
We conduct extensive experiments on ShapeNet-55 with its 34 seen/21 unseen category split to evaluate model performance and generalization capability.

\noindent\textbf{Quantitative Analysis.}
The quantitative results are presented in Table~\ref{tab:unified_shapenet_completion_results}. 
On the full \textbf{ShapeNet-55} benchmark, Simba demonstrates strong performance, achieving a competitive average L2-CD of \textbf{0.79} ($\times 10^3$) and outperforming most prior works. 
This validates the effectiveness of our approach on a wider variety of object categories. More importantly, the results on the \textbf{ShapeNet-34/21} split highlight the superior generalization ability of our framework. 
Our model not only achieves state-of-the-art performance on the 34 seen categories but also maintains a strong lead on the 21 unseen categories. 
This indicates our learned generative prior is more robust and generalizable than deterministic regression, enabling better performance on novel object classes.

\subsection{Results on Real-World Data}

\noindent\textbf{Quantitative and Qualitative Analysis on KITTI.}
To assess real-world performance, we evaluate on the KITTI dataset~\cite{kitti}, which contains sparse and noisy LiDAR scans.
This serves as a challenging domain generalization task, as our model is trained solely on synthetic data.
As reported in Table~\ref{tab:kitti_results}, \textbf{Simba} achieves highly competitive performance, highlighting the robustness of our transformation-based generative approach.
Qualitative results in Figure~\ref{fig:kitti_visuals} further show that our method produces structurally plausible vehicle shapes, avoiding the floating artifacts and scale inconsistencies that plague many competing approaches when facing a significant domain gap.

\subsection{Ablation Studies and Analysis}

We conduct ablation studies on the PCN dataset to analyze our three core components: the diffusion-based transformation prediction, the progressive upsampling strategy, and the heterogeneous MBA-Refiner architecture.

\begin{figure}[t!]
    \centering
    \includegraphics[width=0.9\linewidth]{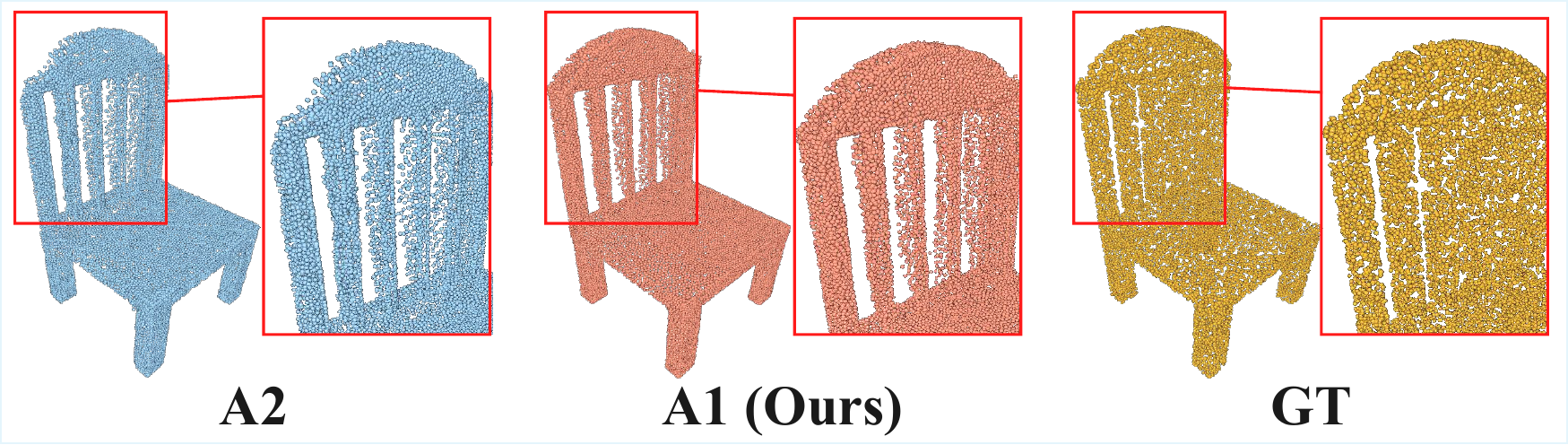}
    \caption{
       Ablation study on the prediction module (A).
    }
    \label{fig:Visual Ablation On A}
\end{figure}

\noindent\textbf{Analysis of the Transformation Prediction Module.}
As shown in Table~\ref{tab:Ablation Study On A and B}, our diffusion-based transformation prediction (A1) clearly outperforms a conventional Transformer regressor (A2), achieving a superior CD score of 6.34 versus 6.48. The visual results in Figure~\ref{fig:Visual Ablation On A} further reinforce this finding; our model reconstructs a geometrically coherent shape, while the regressor produces noticeable and distracting structural artifacts. Collectively, these results validate that learning a generative distribution of transformations is fundamentally more robust, overcoming the critical overfitting to which direct deterministic regression is prone.

\begin{table}[t!]
\centering
\small
\renewcommand{\arraystretch}{1.15}
\setlength{\tabcolsep}{10pt}
\begin{tabular}{l|l|c}
\toprule
\textbf{ID} & \textbf{Configuration} & \textbf{CD-$\ell_1$} ($\times 10^3$) $\downarrow$ \\
\midrule
\textbf{A1} & \textbf{Diffusion Model (Ours)} & \textbf{6.34} \\
A2 & Transformer Regression & 6.48 \\
\midrule
\textbf{B1} & \textbf{3-layer: [2×, 2×, 4×]} & \textbf{6.34} \\
B2 & 1-layer: [16×] & 6.70 \\
B3 & 2-layer: [2×, 8×] & 6.56 \\
B4 & 2-layer: [4×, 4×] & 6.52 \\
\bottomrule
\end{tabular}
\caption{Ablation study on PCN: (A) prediction module and (B) progressive upsampling strategy. All upsampling variants share a total $16\times$ upsampling factor.}
\label{tab:Ablation Study On A and B}
\end{table}

\noindent\textbf{Analysis of the Progressive Upsampling Strategy}
We analyze our progressive upsampling strategy in Table~\ref{tab:Ablation Study On A and B}, comparing different structures that all yield a total $16\times$ upsampling factor. Our three-level upsampling (\textbf{B1}) with a gradual $[2\times, 2\times, 4\times]$ schedule achieves the best performance (6.34 CD-L1). In contrast, a single-level, aggressive $16\times$ upsampling (\textbf{B2}) performs worst (6.70), highlighting the difficulty of direct coarse-to-fine mapping. The two-level variants confirm this trend: the balanced $[4\times, 4\times]$ schedule (\textbf{B4}) significantly outperforms the unbalanced $[2\times, 8\times]$ strategy (\textbf{B3}). This demonstrates that a gradual, multi-level refinement is crucial, and our three-level upsampling approach provides an optimal structure for learning complex geometries.

\begin{figure}[t!]
    \centering
    \includegraphics[width=0.95\linewidth]{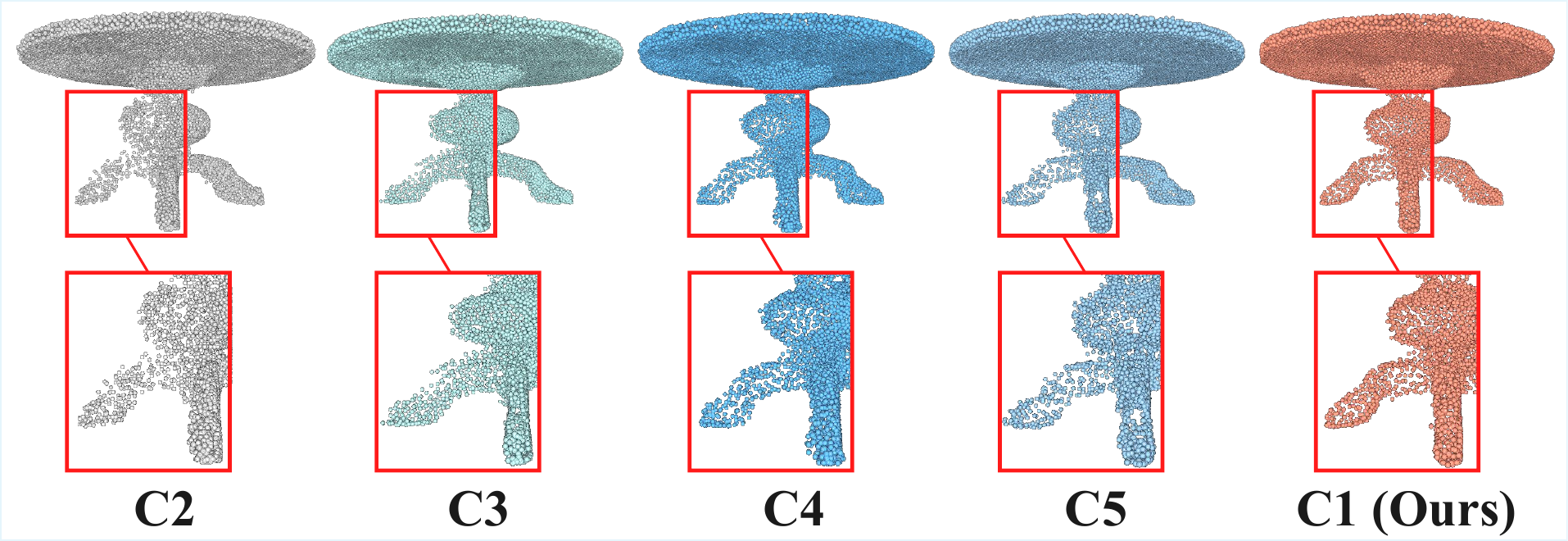}
    \caption{Qualitative comparison of fusion strategies (C).}
    \label{fig:Visual Ablation Study On C}
\end{figure}

\noindent\textbf{Analysis of the MBA-Refiner Architecture.}
We ablate our MBA-Refiner to validate its heterogeneous cascade design. As shown in Table~\ref{tab:Ablation Study On C}, our full model (\textbf{C1}), which uses Cross-Attention in early stages and Mamba in the final stage, sets the performance benchmark with a Chamfer Distance of \textbf{6.34}. First, we test homogeneous architectures. A cascade using only Cross-Attention (\textbf{C4}) exhibits high memory consumption while achieving a slightly inferior score of \textbf{6.35}, demonstrating computational inefficiency. Conversely, using only Mamba blocks (\textbf{C5}) is more memory-efficient but results in a higher error of \textbf{6.43}, proving Cross-Attention is vital for initial feature processing. We then assess simpler fusion methods. Replacing our sequence-based fusion with an MLP-based approach (\textbf{C2}) or using an MLP in the final block (\textbf{C3}) also harms performance, yielding inferior scores of \textbf{6.49} and \textbf{6.41} respectively, despite their lower memory footprints. Figure~\ref{fig:Visual Ablation Study On C} visually confirms these findings, showing that both the MLP-based (\textbf{C2}) and all-Mamba (\textbf{C5}) variants generate distorted geometries, unlike the high-fidelity structure recovered by our model. These results confirm that our heterogeneous cascade (\textbf{C1}) strikes an optimal balance between performance and computational efficiency.

\begin{table}
\centering
\small
\begin{tabular}{l|l|c|c}
\toprule
\textbf{ID} & \textbf{Fusion Strategy per Stage} & \textbf{Memory} & \textbf{CD-$\ell_1$}\\ 
\midrule
\textbf{C1} & \textbf{[CA, CA, MFusion]} (Ours)         & 14.7 GB       & \textbf{6.34} \\
\midrule
C2 & [MLP, MLP, MFusion]                  & 12.1 GB                   & 6.49 \\
C3 & [CA, CA, MLP]                        & \textbf{12.0 GB}         & 6.41 \\
C4 & [CA, CA, CA]                         & 16.4 GB                  & 6.35 \\
C5 & [MFusion, MFusion, MFusion]          &  13.8 GB                & 6.43 \\
\bottomrule
\end{tabular}
\caption{Ablation study on PCN for MBA-Refiner fusion strategies (C). Different fusion combinations are compared using CD-$\ell_1$ scores ($\times 10^3$). CA denotes Cross-Attention and MFusion denotes Mamba Fusion.}
\label{tab:Ablation Study On C}
\end{table}

\section{Conclusion}
We introduce Simba, a novel paradigm for point cloud completion which reformulates the task as learning the distribution of geometric transformations via diffusion. This is achieved via two core components: (1) a Symmetry-Diffusion mechanism (Sym-Diffuser) to address the overfitting and noise-sensitivity issues of prior direct-regression methods, and (2) a cascaded Mamba-based architecture (MBA-Refiner) for high-fidelity upsampling. Extensive experiments demonstrate that Simba learns robust transformation representations, enabling strong cross-domain generalizability on the real-world KITTI benchmark, achieving state-of-the-art performance on multiple benchmarks. Consequently, Simba establishes a new direction for diffusion-based point cloud completion.
\newpage
\section{Acknowledgements}
This work was supported in part by the Natural Science
Foundation of China (No. 62272227 and No. U25A20533).
\bibliography{aaai2026}
\makeatletter
\makeatother

\clearpage
\appendix
\section{Preliminaries}

Our work builds upon the framework of Denoising Diffusion Probabilistic Models (DMs)~\cite{ddpm, ddim}. DMs consist of a forward noising process and a reverse denoising process.

\subsection{Forward Diffusion Process}
The forward process gradually adds Gaussian noise to a clean data sample $\boldsymbol{\mathcal{Z}}_0$ over $T$ discrete timesteps. In our work, $\boldsymbol{\mathcal{Z}}_0$ represents the ground truth transformation field. This process is defined as a Markov chain:
\begin{equation}
q(\boldsymbol{\mathcal{Z}}_t | \boldsymbol{\mathcal{Z}}_{t-1}) = \mathcal{N}(\boldsymbol{\mathcal{Z}}_t; \sqrt{1 - \beta_t} \boldsymbol{\mathcal{Z}}_{t-1}, \beta_t \mathbf{I})
\end{equation}
where $\beta_t \in (0, 1)$ is a predefined variance schedule. A key property of this process is that we can sample the noisy field $\boldsymbol{\mathcal{Z}}_t$ at any arbitrary timestep $t$ directly from the initial clean field $\boldsymbol{\mathcal{Z}}_0$ in a closed form:
\begin{equation}
q(\boldsymbol{\mathcal{Z}}_t | \boldsymbol{\mathcal{Z}}_0) = \mathcal{N}(\boldsymbol{\mathcal{Z}}_t; \sqrt{\bar{\alpha}_t} \boldsymbol{\mathcal{Z}}_0, (1 - \bar{\alpha}_t)\mathbf{I})
\end{equation}
where $\alpha_t = 1 - \beta_t$ and $\bar{\alpha}_t = \prod_{i=1}^{t} \alpha_i$.
This allows us to express $\boldsymbol{\mathcal{Z}}_t$ using a reparameterization trick with a standard Gaussian noise $\boldsymbol{\epsilon} \sim \mathcal{N}(\mathbf{0}, \mathbf{I})$:
\begin{equation}
\boldsymbol{\mathcal{Z}}_t = \sqrt{\bar{\alpha}_t} \boldsymbol{\mathcal{Z}}_0 + \sqrt{1 - \bar{\alpha}_t} \boldsymbol{\epsilon}.
\label{eq:supp_forward_reparam}
\end{equation}
\subsection{Reverse Denoising Process}
The reverse process aims to learn a neural network to denoise $\boldsymbol{\mathcal{Z}}_t$ back to $\boldsymbol{\mathcal{Z}}_{t-1}$, conditioned on external information. The posterior distribution $p(\boldsymbol{\mathcal{Z}}_{t-1}|\boldsymbol{\mathcal{Z}}_t, \boldsymbol{\mathcal{Z}}_0)$ is tractable and given by:
\begin{equation}
p(\boldsymbol{\mathcal{Z}}_{t-1} | \boldsymbol{\mathcal{Z}}_t, \boldsymbol{\mathcal{Z}}_0) = \mathcal{N}(\boldsymbol{\mathcal{Z}}_{t-1}; \boldsymbol{\mu}_t(\boldsymbol{\mathcal{Z}}_t, \boldsymbol{\mathcal{Z}}_0), \sigma_t^2 \mathbf{I})
\end{equation}
where $\boldsymbol{\mu}_t(\boldsymbol{\mathcal{Z}}_t, \boldsymbol{\mathcal{Z}}_0) = \frac{1}{\sqrt{\alpha_t}}(\boldsymbol{\mathcal{Z}}_t - \frac{1 - \alpha_t}{\sqrt{1 - \bar{\alpha}_t}} \boldsymbol{\epsilon})$ and $\sigma_t^2$ are variance terms derived from $\beta_t$.

Since the ground truth $\boldsymbol{\mathcal{Z}}_0$ is unknown during inference, the model must learn to approximate this reverse transition. Our model, Simba, is a \textbf{conditional} diffusion model that operates as a noise predictor. It takes the noisy field $\boldsymbol{\mathcal{Z}}_t$, the timestep $t$, and a conditioning feature vector $\boldsymbol{\mathcal{F}}_k$ (extracted from the partial input) to predict the noise component, denoted as $\boldsymbol{\epsilon}_\theta(\boldsymbol{\mathcal{Z}}_t, t, \boldsymbol{\mathcal{F}}_k)$. The one-step denoising transition can then be formulated as:
\begin{equation}
\boldsymbol{\mathcal{Z}}_{t-1} = \frac{1}{\sqrt{\alpha_t}}\left(\boldsymbol{\mathcal{Z}}_t - \frac{1 - \alpha_t}{\sqrt{1 - \bar{\alpha}_t}} \boldsymbol{\epsilon}_\theta(\boldsymbol{\mathcal{Z}}_t, t, \boldsymbol{\mathcal{F}}_k)\right) + \sigma_t \mathbf{z},
\end{equation}
where $\mathbf{z} \sim \mathcal{N}(\mathbf{0}, \mathbf{I})$ is newly sampled noise.

To connect this process to the objective function in the main paper, we clarify the role of the noise predictor $\boldsymbol{\epsilon}_\theta(\boldsymbol{\mathcal{Z}}_t, t, \boldsymbol{\mathcal{F}}_k)$. While the network is trained to directly predict the noise component, our training objective, the proxy loss, is formulated based on the predicted clean data. This predicted clean transformation field, denoted as $\hat{\boldsymbol{\mathcal{T}}}_{\theta}$. It represents an estimate of the original clean data $\boldsymbol{\mathcal{Z}}_0$, which is derived by algebraically inverting the forward diffusion step (from Eq. (10)) using the predicted noise:
\begin{equation}
\hat{\boldsymbol{\mathcal{T}}}_{\theta}(\boldsymbol{\mathcal{Z}}_t, t, \boldsymbol{\mathcal{F}}_k) = \frac{1}{\sqrt{\bar{\alpha}_t}} \left(\boldsymbol{\mathcal{Z}}_t - \sqrt{1 - \bar{\alpha}_t} \boldsymbol{\epsilon}_\theta(\boldsymbol{\mathcal{Z}}_t, t, \boldsymbol{\mathcal{F}}_k)\right).
\label{eq:supp_pred_clean}
\end{equation}
In practice, the expectation over all timesteps in the proxy loss, defined in Eq.~\ref{eq:proxy_loss}, is approximated by a discrete summation. Specifically, we evaluate the loss at a pre-selected, fixed set of $K=10$ timesteps for each training instance. The final proxy loss is then computed as the weighted sum of the Mean Squared Errors between the predicted clean transformation field $\hat{\boldsymbol{\mathcal{T}}}_{\theta}$ and the ground-truth field $\boldsymbol{\mathcal{T}}_{\text{gt}}$ at these specific timesteps.

\section{Model Analysis}

We analyze Simba's efficiency by comparing its parameter count and performance against several methods on the PCN benchmark. As shown in Table~\ref{tab:params_perf_transposed}, Simba achieves a state-of-the-art  CD-$\ell_1$ score with a moderate parameter count. This result highlights its architectural efficiency, striking a compelling balance between computational complexity and reconstruction fidelity.

\begin{table}[h]
\centering
\small

\begin{tabular}{lcc}
\toprule
\textbf{Model} & \textbf{Parameters (M)} &   CD-$\ell_1$ ($\times 10^3$) $\downarrow$  \\
\midrule
PCN & \textbf{6.55} & 9.64 \\
SymComplete & 13.28 & 6.47 \\
SnowFlakeNet & 19.29 & 7.21 \\
AdaPoinTr & 32.49 & 6.53 \\
PointCFormer & 33.20 & 6.41 \\
\textbf{Simba (Ours)} & 23.66 & \textbf{6.34} \\
\bottomrule
\end{tabular}
\caption{Comparison of model parameters and CD-$\ell_1$ performance on the PCN dataset. Lower CD-$\ell_1$ is better.}
\label{tab:params_perf_transposed}
\end{table}

\section{Training Details}
Across all datasets and training stages, we employ a unified optimization strategy. Models are trained using the AdamW optimizer with a weight decay of $5 \times 10^{-4}$. The learning rate is governed by a cosine annealing scheduler featuring a 20-epoch warmup, where the rate peaks at $2 \times 10^{-4}$ and subsequently decays to a minimum of $1 \times 10^{-5}$. Our transformation diffusion module is based on a 100-step DDPM formulation; for generating the transformation field, we consistently use an accelerated 25-step DDIM sampling process during both training and inference.

\section{More Visualization}
In this supplementary material, we provide extensive qualitative results to further demonstrate the effectiveness and robustness of our proposed Simba framework. We showcase Simba's capability to complete asymmetrical objects, providing evidence that our method is not constrained by simplistic global symmetry assumptions. Additionally, we present more visual comparisons on the PCN~\cite{pcn} and ShapeNet-55~\cite{shapenet} datasets to highlight its superior performance in generating high-fidelity and geometrically consistent completions across various object categories. Finally, we include more examples from the real-world KITTI~\cite{kitti} dataset to underscore its strong cross-domain generalization from synthetic training data to sparse and noisy LiDAR scans. These results collectively reinforce the claims made in the main paper.


\subsection{Handling Asymmetrical Objects}
\begin{figure*}
    \centering
    \includegraphics[width=0.9\linewidth]{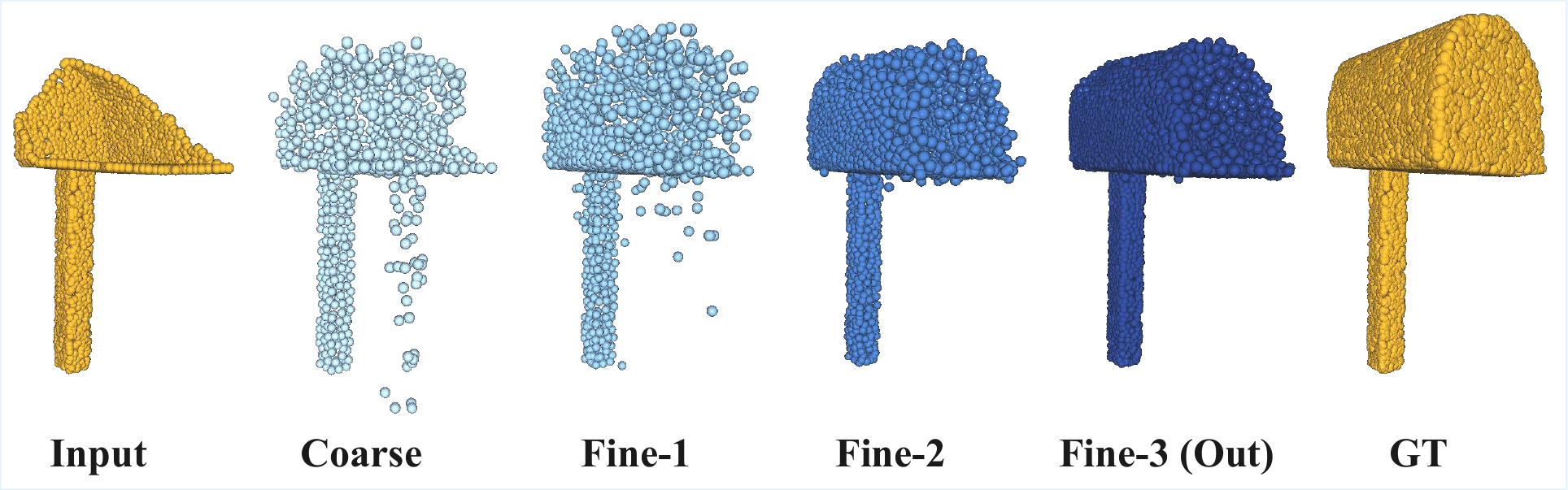}
    \caption{Qualitative results on an asymmetrical 'mailbox' object. The figure demonstrates the progressive refinement from a coarse shape to the final output (Fine-3), showing that Simba can handle non-symmetrical structures effectively.}
    \label{fig:appendix_asymmetrical_object}
\end{figure*}

A key limitation of many symmetry-based methods is their struggle with asymmetrical objects, often imposing a false symmetry that distorts the result. As shown in Figure~\ref{fig:appendix_asymmetrical_object}, our Simba framework addresses this challenge. While our initial coarse completion provides a structurally sound base, it may temporarily enforce symmetry, as seen with the mirrored flap in the 'Coarse' stage. The key to handling asymmetry is our cascaded refinement module, which progressively corrects these artifacts by removing the unwanted mirrored geometry. This demonstrates that combining a symmetry-aware initial proposal with a strong refinement process is crucial for accurately completing asymmetrical objects.

\subsection{Additional Qualitative Results on the PCN Dataset}
\begin{figure*}
    \centering
    \includegraphics[width=0.9\linewidth]{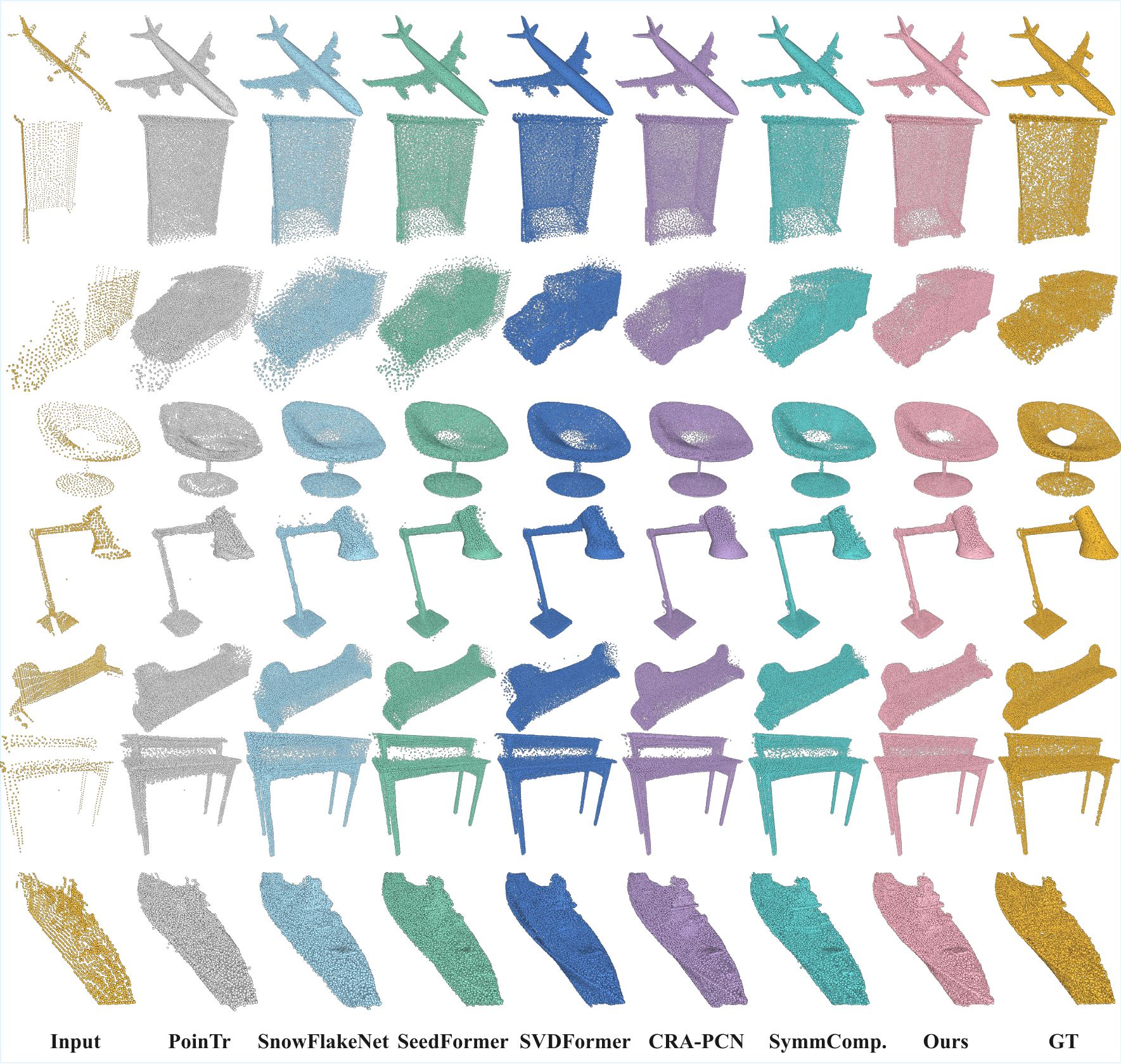}
    \caption{More qualitative comparisons on the PCN dataset. Our method (Ours) consistently generates completions that are both structurally sound and detailed, outperforming competing methods across various categories.}
    \label{fig:appendix_pcn}
\end{figure*}

In contrast to existing approaches that struggle with structural coherence, Simba demonstrates superior geometric consistency as highlighted in Figure~\ref{fig:appendix_pcn}. While many prior methods produce fragmented artifacts (e.g., PoinTr, SnowFlakeNet) or distorted shapes (e.g., CRA-PCN), Simba consistently reconstructs objects with coherent global structures and preserved fine details. This is evident in the smooth surfaces of the airplane and the slender, continuous legs of the chair and table. These results validate that our transformation-based diffusion approach effectively enforces structural integrity, leading to more plausible and robust completions.

\subsection{Additional Qualitative Results on the ShapeNet-55 Dataset}
\begin{figure*}
    \centering
    \includegraphics[width=0.9\linewidth]{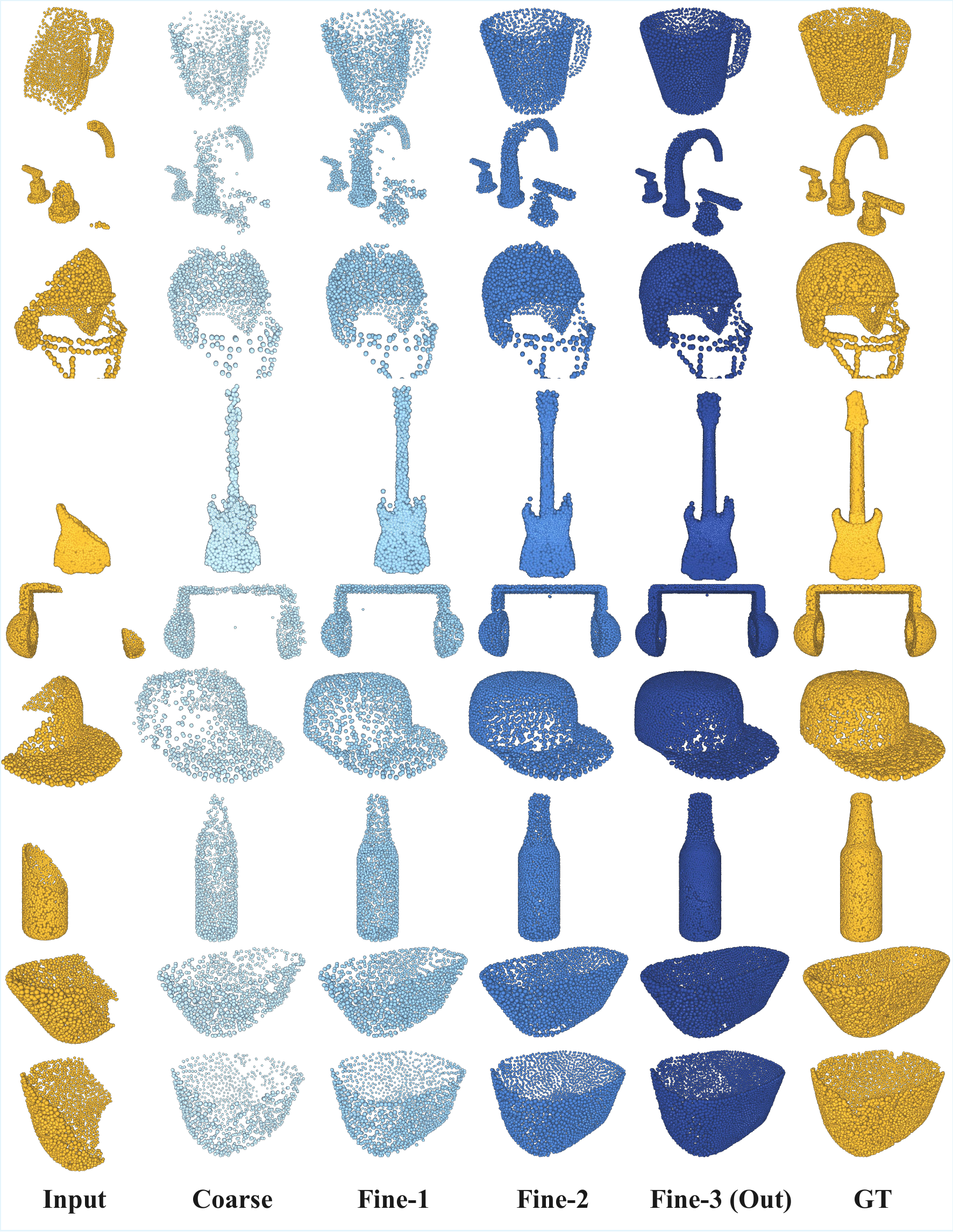}
    \caption{More qualitative results on the diverse ShapeNet-55 dataset. The figure illustrates the intermediate steps of our refinement process, showing robust performance on complex objects such as a mug, faucet, helmet, guitar, headphones, hat, bottle, and bowl.}
    \label{fig:appendix_shapenet}
\end{figure*}

To showcase the generalization capability of our model on more diverse and complex shapes, we present additional results on the ShapeNet-55 dataset in Figure~\ref{fig:appendix_shapenet}. Simba demonstrates robust performance across a wide range of object categories, successfully reconstructing intricate structures (like the spout of a faucet, the frame of a helmet, or the body of a guitar) while maintaining global coherence. This highlights the effectiveness of our transformation diffusion and Mamba-based refiner in handling varied geometries.

\subsection{Additional Qualitative Results on the KITTI Dataset}
\begin{figure*}
    \centering
    \includegraphics[width=0.83\linewidth]{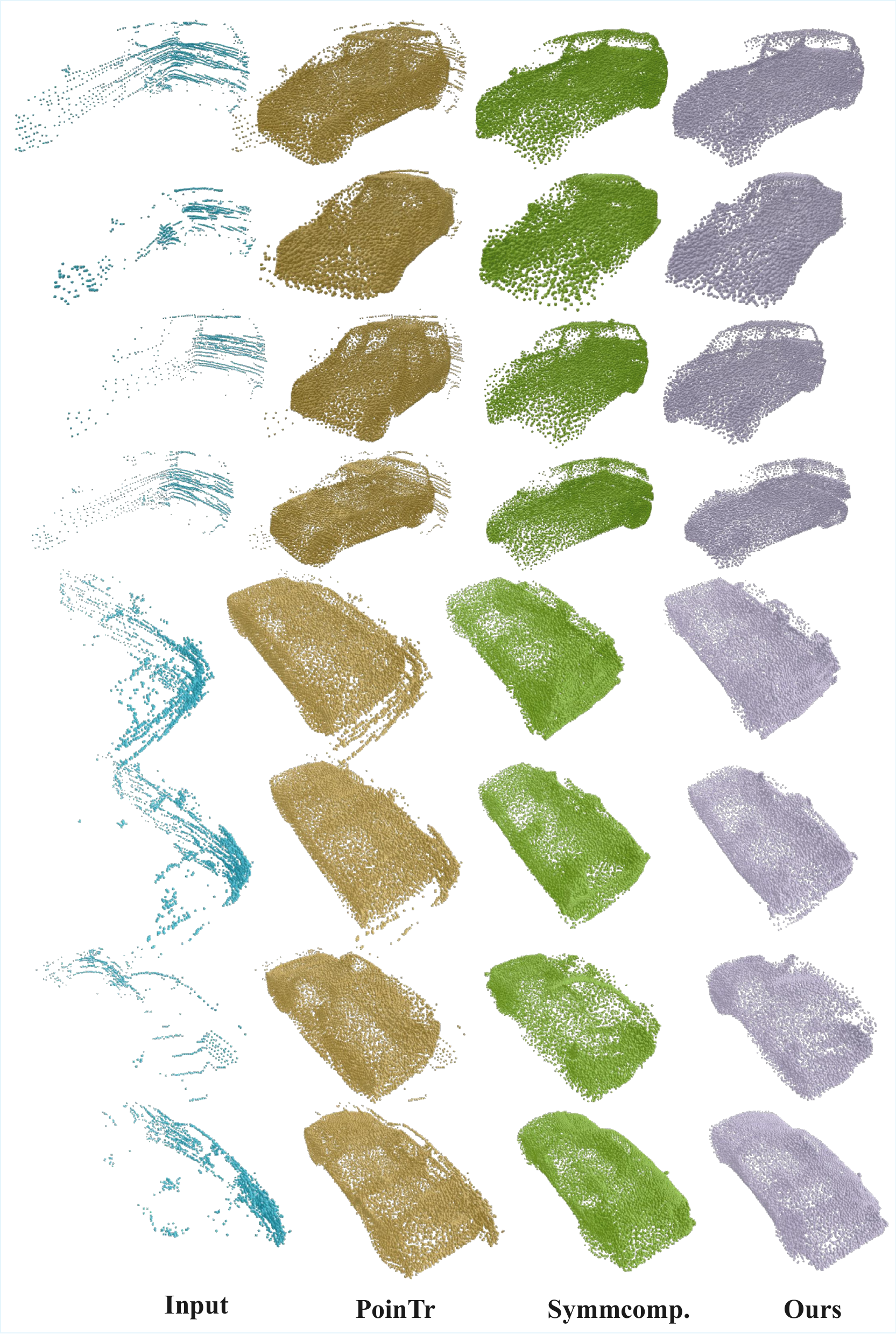}
    \caption{More qualitative comparison of cross-domain completion results on KITTI dataset.}
    \label{fig:appendix_kitti}
\end{figure*}

The challenging conditions present in real-world LiDAR data further validate our approach's effectiveness, with additional qualitative results on the KITTI dataset presented in Figure~\ref{fig:appendix_kitti}. When faced with highly sparse and fragmented real-world LiDAR scans, competing methods often produce amorphous or incomplete results. In contrast, Simba consistently generates coherent and structurally complete vehicle shapes, faithfully recovering the overall form. This visual evidence highlights the effectiveness of our transformation-based generative approach in bridging the synthetic-to-real domain gap and handling challenging, noisy data.

\end{document}